\documentclass[lettersize,journal]{IEEEtran}
\usepackage{amsmath,amsfonts}
\usepackage{diagbox}
\usepackage{array}
\usepackage{makecell} 
\usepackage{booktabs}
\usepackage{diagbox} 
\usepackage{subcaption}
\usepackage{textcomp}
\usepackage{stfloats}
\usepackage{url}
\usepackage{hyperref}
\usepackage{verbatim}
\usepackage{graphicx}
\usepackage{cite}
\usepackage{arydshln}
\usepackage{makecell}
\usepackage{arydshln}
\usepackage{booktabs}
\usepackage{placeins}

\usepackage{xcolor}
\usepackage{colortbl}
\usepackage{algorithm}
\usepackage{algpseudocode}
\usepackage{tikz}
\usepackage{mathrsfs} 
\usepackage{amsmath}
\usepackage[most]{tcolorbox}
\newtcolorbox{blackalgobox}[1][]{
  breakable,
  colback=white,
  colframe=black,    
  title=#1,
  fonttitle=\bfseries,
  sharp corners,
  boxrule=0.8pt,
}

\definecolor{blue}{rgb}{0, 0.5, 1}  
\usepackage[font=small]{caption}
\begin{document}

\title{Unified Multimodal Vessel Trajectory Prediction with Explainable Navigation Intention} 

\author{Rui Zhang, Chao Li, Kezhong Liu, Chen Wang ~\IEEEmembership{Senior Member,~IEEE,} Bolong Zheng ~\IEEEmembership{Member,~IEEE,} \\and Hongbo Jiang ~\IEEEmembership{Senior Member,~IEEE} 

\thanks{Manuscript received XX XX, 2024.~
This work was supported in part by the National Natural Science Foundation of China under Grant 52031009, Grant 62372161, Grant 62272183, and Grant 62372194; in part by the Yuelushan Center for Industrial Innovation under Grant 2025YCII0127 and  in part the Key R\&D Program of Hubei Province under Grant 2025EHA033.}

~\textit{(The corresponding authors of this paper is K.~Liu.)}
\thanks{Rui Zhang, Chao Li and Bolong Zheng are with the School of Computer Science and Artificial Intelligence, Wuhan University of Technology, Wuhan, 430000, China. E-mail: \{zhangrui, 302476, bolongzheng\}@whut.edu.cn.}  
\thanks{Kezhong Liu is with the School of Navigation, Wuhan University of Technology, Wuhan, 430000, China. E-mail: kzliu@whut.edu.cn.} 
\thanks{Chen Wang is with the Hubei Key Laboratory of Internet of Intelligence, School of Electronic Information and Communications, Huazhong University of Science and Technology, Wuhan 430074, China. E-mail: cwangwhu@gmail.com}  
\thanks{Hongbo Jiang is with the College of Computer Science and Electronic Engineering, Hunan University, Changsha 410012, China. E-mail: hongbojiang2004@gmail.com}
}  

\markboth{Journal of \LaTeX\ Class Files,~Vol.~14, No.~8, August~2021}%
{Shell \MakeLowercase{\textit{et al.}}: A Sample Article Using IEEEtran.cls for IEEE Journals}
\IEEEaftertitletext{\vspace{-2\baselineskip}}
\maketitle

\begin{abstract}  
Vessel trajectory prediction is fundamental to intelligent maritime systems. Within this domain, short-term prediction of rapid behavioral changes in complex maritime environments has established multimodal trajectory prediction (MTP) as a promising research area. However, existing vessel MTP methods suffer from limited scenario applicability and insufficient explainability. To address these challenges, we propose a unified MTP framework incorporating explainable navigation intentions, which we classify into sustained and transient categories. Our method constructs sustained intention trees from historical trajectories and models dynamic transient intentions using a Conditional Variational Autoencoder (CVAE), while using a non-local attention mechanism to maintain global scenario consistency. Experiments on real Automatic Identification System (AIS) datasets demonstrates our method's broad applicability across diverse scenarios, achieving significant improvements in both ADE and FDE. Furthermore, our method improves explainability by explicitly revealing the navigational intentions underlying each predicted trajectory. 
\end{abstract}  

\begin{IEEEkeywords}
Vessel Trajectory, Multimodal Trajectory Prediction, Explainability
\end{IEEEkeywords}

\section{Introduction}
\IEEEPARstart{A}{rtificial} intelligence has transformed maritime transportation, with trajectory prediction emerging as a foundation for intelligent vessel development. This task, which predicts navigation paths from observed trajectories, supports critical maritime operations including path planning\cite{ref1}, collision avoidance\cite{ref2}, and traffic flow management\cite{ref3}. 

Vessel trajectory prediction includes two temporal scales. Long-term prediction\cite{ref4,ref5}, spanning hours to days, relies on deterministic trajectory prediction (DTP) since vessels follow standardized routes with minimal deviations. 

Short-term prediction\cite{ref6,ref7}, operating on minute-level intervals, requires precise modeling of rapid trajectory changes within complex maritime environments. Its critical role in navigation safety, port operations, and emergency response has established it as the field's primary focus. DTP's inherent limitations in capturing dynamic behaviors have necessitated the development of multimodal trajectory prediction (MTP)\cite{ref7,ref8}, which generates multiple possible routes\cite{ref9} by incorporating historical data and environmental context.

Current MTP methods face significant limitations. Certain methods~\cite{ref7} restrict themselves to two-vessel encounters, while others~\cite{ref8} disregard encounter dynamics entirely. Both methods mainly model uncertainty around a single dominant modal, rather than capturing multiple future modals. As the result, they assign uniform probabilities across predicted outputs and lack sufficient explainability.

\setlength{\belowcaptionskip}{4pt}
\begin{figure*}[!t]   
\centering  
\includegraphics[width=\textwidth]{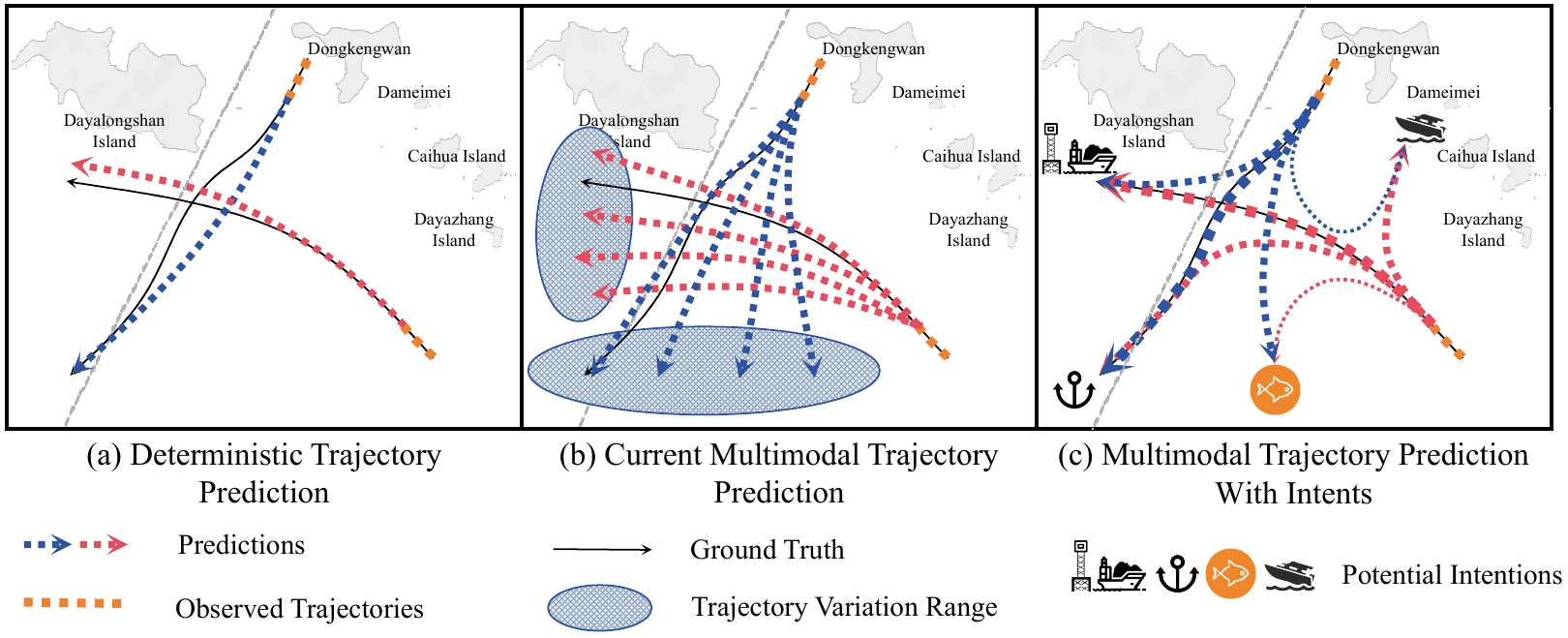}  
\caption{Comparison of vessel trajectory prediction methods, where orange dashed lines indicate observed trajectories, black solid lines show actual trajectories, and blue/red dashed lines represent predicted trajectories for two vessels. (a) Deterministic prediction generates a single fixed trajectory, emphasizing route selection while disregarding vessel encounters. (b) Current multimodal prediction, shown with a light blue area indicating trajectory variation range, achieves predictions with equal probabilities and limited modal diversity. (c) Our proposed multimodal intention-aware prediction offers diverse, realistic trajectories with differentiated probabilities (indicated by line thickness) and explicit navigation intentions (represented by icons).} 
\label{fig_sim}  
\end{figure*}

Fig.~\ref{fig_sim} illustrates the progression from deterministic to multimodal prediction. While deterministic methods focus solely on route selection\cite{ref27,ref11}, existing MTP methods, though generating multiple trajectories, exhibit limited diversity and assume uniform probability distributions\cite{ref7,ref8}.

We propose DI-MTP (Dual Intention-Multimodal Trajectory Prediction), a unified framework that distinguishes between sustained intentions (stable patterns toward fixed targets) and transient intentions (real-time environmental adaptations). This dual-intention method improves prediction accuracy and explainability across diverse scenarios by generating multimodal predictions with associated probability estimates.

Implementation of this framework presents two key challenges: (1) extracting sustained intentions from incomplete trajectories while accommodating varied vessel behaviors across different contexts, and (2) modeling complex sustained-transient intention interactions influenced by diverse interpretations of COLREGS (International Regulations for Preventing Collisions at Sea)\cite{ref10} and vessel states.

To address these challenges, we introduce sustained intention tree construction and global transient intention optimization. Our method integrates Conditional Variational Autoencoder (CVAE)-based intention modeling with non-local attention mechanisms to ensure consistent scenario-wide predictions.
  
Our main contributions are as follows:  
  
\begin{enumerate}  
\item We present the first unified multimodal vessel trajectory prediction model (DI-MTP) that generalizes across diverse scenarios, including both non-encounter and multi-vessel encounter situations.
\item We propose an explainable framework leveraging attention-based intention trees to provide modal probabilities for multiple predictions, enabling explainability through distinct sustained intention prototypes.
\item Extensive evaluation on real AIS datasets demonstrating significant improvements over state-of-the-art methods: 18.28\% in ADE and 8.51\% in FDE within the Zhoushan dataset and 27.48\% in ADE and 27.40\% in FDE within the Hainan dataset.  The code of DI-MTP is available at: \url{https://github.com/EurusC/DI-MTP}.
\end{enumerate}  
  
The rest of this paper is organized as follows. Section II reviews related work, Sections III and IV present preliminaries and method details, Section V discusses experiments, and Section VI concludes.

\section{Related Work}  
Vessel trajectory prediction operates in a complex dynamic environment where vessel behaviors are influenced by multi-source heterogeneous factors, including rapidly changing meteorological conditions, hydrodynamic interactions, multi-vessel collision-avoidance strategies, and sensor noise from heterogeneous navigation systems (e.g., AIS, radar, and surveillance video streams) \cite{ref33}, \cite{ref34}. 
This complexity inherently induces multimodal characteristics in vessel trajectory predictions.

\subsection{Deterministic Vessel Trajectory Prediction}  

DTP methods are the foundation of vessel trajectory prediction, leveraging deep learning methods to process complex time-series data and model long-term vessel motion patterns.

Long Short-Term Memory (LSTM) networks have demonstrated effectiveness in modeling long-term dependencies. Gao et al.\cite{ref11} combined trajectory similarity analysis with LSTM to generate navigation support points. Liu et al.\cite{ref12} integrated quaternion vessel domains with LSTM to capture both dynamic and static vessel characteristics. Zaman et al.\cite{ref39} comprehensively compared LSTM, GRU, DBLSTM, and Kalman Filter models. In a follow-up study\cite{ref40}, they demonstrated that CNN-based methods perform well when applied to clean AIS data, highlighting the importance of data quality in trajectory prediction.

Encoder-Decoder architectures further improved DTP capabilities. Nguyen et al.\cite{ref5} developed a grid-based LSTM encoder-decoder for trajectory prediction, while Capobianco et al.\cite{ref13} implemented a sequence-to-sequence model using bidirectional LSTM encoders and unidirectional decoders.

Attention mechanisms and Transformers improved long-term dependency modeling. Sekhon et al.\cite{ref14} introduced an attention-based bidirectional LSTM with dynamic temporal weighting. Wang et al.\cite{ref15} and Jiang et al.\cite{ref16} used Transformers for processing AIS spatio-temporal features. Liu et al.\cite{ref6} combined graph convolutional networks with Transformers, using multi-graph representations and mobile edge computing for optimization.  
  
\subsection{Multimodal Vessel Trajectory Prediction}  

MTP generates multiple possible trajectories per vessel. Murray and Perera\cite{ref17} used Gaussian Mixture Models for route clustering, though their single-point neighborhood approach showed limitations in prediction and inter-vessel interaction modeling.

Generative models improved MTP capabilities. Guo et al.\cite{ref8} implemented GAN-based reversible mapping between latent vectors and trajectories, improving prediction diversity but struggling with multi-vessel scenarios. Han et al.\cite{ref7} applied CVAEs to ferry-merchant encounters but failed to generalize or explain prediction multimodality.

Our method addresses these limitations through a comprehensive MTP model integrating navigation intentions, multimodality, and inter-vessel interactions. 

\section{Preliminary}
Before delving into our model, we first present the fundamental elements of our research, including key definitions, data preprocessing workflow, and underlying assumptions.

\subsection{Problem Formulation}
\label{problem formulation}
\textbf{Definition 1 (Trajectory $\mathcal{T}$):} A trajectory $\mathcal{T}$ represents an ordered sequence of vessel positions, defined as $\mathcal{T} = \{(x_1, y_1), (x_2, y_2), \ldots, (x_L, y_L)\}$, where $L$ is the sequence length, and $(x_i, y_i)$ is the vessel's longitude and latitude at timestep $i$.

\textbf{Definition 2 (Scenario $\mathcal{S}$):} A scenario $\mathcal{S}$ is a set of $m$ trajectories occurring within the same spatiotemporal domain, denoted as $\mathcal{S} = \{\mathcal{T}_1, \mathcal{T}_2, \ldots, \mathcal{T}_m\}$, where $\mathcal{T}_i$ is the $i$-th trajectory within $\mathcal{S}$. We denote the collection of prior scenarios available during the training phase as $\mathcal{S}^h$.

\textbf{Definition 3 (Multimodality $k$ and $n$):}
\label{Definition3}

We characterize multimodality by $k$ distinct sustained intentions, and generate a total of $n$ predicted trajectories by modeling uncertainty within each sustained intention, where $k \leq n$.

\textbf{Problem Statement:}
\label{problem-statement}

Given a scenario $\mathcal{S}$, we denote the observed and predicted segment of each trajectory $\mathcal{T} \in \mathcal{S}$ as  
\[
\mathcal{T}^o = \{(x_t, y_t)\}_{t = 1}^{L_o}, \quad \mathcal{T}^p = \{(x_t, y_t)\}_{t = L_o+1}^{L_o + L_p}
\]  
where \(\mathcal{T}^o\) corresponds to the first \(L_o\) timesteps, and \(\mathcal{T}^p\) corresponds to the subsequent \(L_p\) timesteps as the ground truth for prediction. Let $\mathcal{S}^o = \{\mathcal{T}^o_1, \mathcal{T}^o_2, \ldots, \mathcal{T}^o_m\}$ and $\mathcal{S}^p = \{\mathcal{T}^p_1, \mathcal{T}^p_2, \ldots, \mathcal{T}^p_m\}$ be the sets of observed and corresponding ground truth future trajectories in the scenario, respectively.

Our objective is to generate the multimodal predictions for each vessel in the scenario over the next $L_p$ timesteps, resulting in $\hat{\mathcal{S}^p} = \{\hat{\mathcal{T}^p_1}, \hat{\mathcal{T}^p_2}, \ldots, \hat{\mathcal{T}^p_m\}}$. Each predicted trajectory $\hat{\mathcal{T}^p_i}$ is a set of $n$ plausible future continuations:  
\[
\hat{\mathcal{T}^p_i} = \{\hat{\mathcal{T}^{p}_{i, 1}}, \hat{\mathcal{T}^{p}_{i, 2}}, \ldots, \hat{\mathcal{T}^{p}_{i, n}\}},
\]  
and, each candidate $\hat{\mathcal{T}^{p}_{i, j}}$ is defined as  
\[
\hat{\mathcal{T}^{p}_{i, j}} = \{(x^{\hat{\mathcal{T}^p_{i,j}}}_t, y^{\hat{\mathcal{T}^p}_{i,j}}_t)\}_{t = L_o+1}^{L_o + L_p},
\]
where $(x^{\hat{\mathcal{T}^p_{i,j}}}_t, y^{\hat{\mathcal{T}^p_{i,j}}}_t)$ denotes the coordinates of trajectory $\hat{\mathcal{T}^p_{i,j}}$ at timestep $t$.

\subsection{Data Preprocessing}
Given the inherent challenges of AIS data, such as missing values, irregular time intervals, and outliers, we implemented a series of preprocessing steps.

\subsubsection{Track Extraction}
AIS messages are grouped by Maritime Mobile Service Identity (MMSI) to extract vessel-specific tracks. Messages are sorted by timestamp, duplicates are removed, and points outside the area of interest are discarded.

\subsubsection{Segmentation \& Filtering}
Each track is segmented into continuous fragments based on temporal gaps. A new segment begins when the time difference between two consecutive points exceeds a threshold (e.g., 10 minutes). Segments with fewer than a minimum number of points (e.g., 20) are discarded.

\subsubsection{Temporal Resampling}
To address irregular time intervals, each segment is resampled at fixed intervals (e.g., 30 seconds) using cubic spline interpolation, resulting in uniformly spaced trajectory points.

\subsubsection{Scenario Construction}
A time window with dense maritime traffic is selected, and all active segments within this period are aggregated to form a scenario $\mathcal{S}$, which contains vessel movements within a shared spatiotemporal context.

\subsubsection{Encounter Mask Generation}
For each scenario, a binary encounter mask $M \in \mathbb{R}^{(m, m)}$ is constructed by evaluating all trajectory pairs. For a pair $(i, j)$, if both the Time to Closest Point of Approach (TCPA) and Distance to Closest Point of Approach (DCPA) fall within predefined thresholds (e.g., TCPA $\in [-0.3, 0.8]$ hours, DCPA $\in [0, 2]$ nautical miles), the pair is marked as an encounter ($M_{ij} = 1$); otherwise, it is marked as a non-encounter ($M_{ij} = 0$).

\subsection{CVAE and Multimodality}

Multimodality is an inherent characteristic of vessel trajectory prediction, where a single observed trajectory $\mathcal{T}^o$ can correspond to multiple plausible future paths due to different navigational intentions and uncertainties. To formalize this multimodality, we propose these modeling assumptions.

1) We adopt the CVAE framework, which has demonstrated strong capability in capturing complex conditional distributions in numerous trajectory prediction studies \cite{ref7, ref26, ref37, ref38}. CVAE enables diverse predictions by sampling from a latent space conditioned on contextual and interaction information.

2) As stated in Definition 3 in Section~\ref{Definition3}, we decompose trajectory multimodality into two components: (1) distinct sustained intentions leading to discrete modals, and (2) transient intentions within each modal, accounting for fine-scale variations characterized by Gaussian-distributed uncertainties. 

We assume each trajectory stems from sustained and transient intentions. Sustained intentions act as conditions, while transient intentions capture variability as distributions. This structure naturally fits the CVAE framework.

\section{Methodology}  
\subsection{Overview}

\begin{figure*}[!t]  
\centering  
\includegraphics[width=\textwidth]{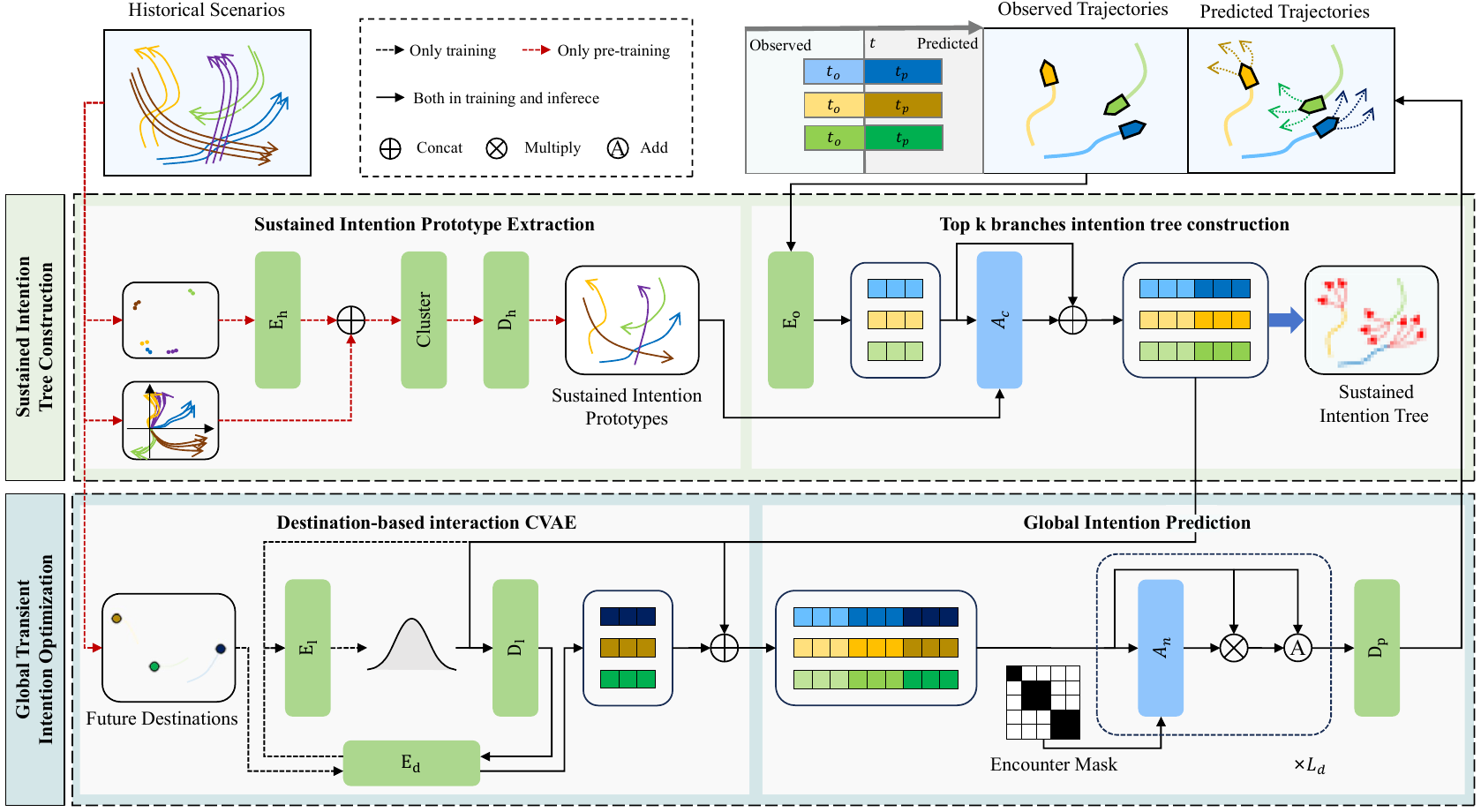}  
\caption{Overall of DI-MTP. Sustained Intentions are extracted from historical trajectories and matched with observed trajectories to construct sustained intention trees. Next, the relationships between sustained and transient intentions are captured by modeling trajectory destinations. The overall encoding passes through a non-local attention mechanism to further strengthen the relationships between vessels in the global context, and finally, the decoding obtains the final prediction.}  
\label{fig_framework}  
\end{figure*}  

As shown in Fig.~\ref{fig_framework}, the DI-MTP framework takes two types of input: (1) \textbf{Historical Scenarios} $\mathcal{S}^h$, representing accessible prior scenarios (e.g., training dataset); (2) \textbf{Observed Scenario} $\mathcal{S}^o$, and outputs the \textbf{Predicted Scenario} $\hat{\mathcal{S}^p}$. These inputs and output are visualized in the top-left and top-right of Fig.~\ref{fig_framework}.

The DI-MTP framework consists of two main components:
\begin{enumerate}
    \item \textbf{Sustained Intention Tree Construction:} Constructs a sustained intention tree by extracting prototypes as priors from $\mathcal{S}^h$ and matching them with the $\mathcal{S}^o$ context.
    \item \textbf{Global Transient Intention Optimization:} Models the dynamic interaction between sustained and transient intentions via destination prediction and performs global trajectory forecasting.
\end{enumerate}

\subsection{Sustained Intention Tree Construction} 
\label{subsec:sustained_intention_tree}
We introduce a sustained intention tree that provides explainable modal sources through: (1) extracting sustained intention prototypes and (2) attention-based intention tree construction. (Algorithm \ref{alg:continued_intention_tree}) 

\textbf{Sustained Intention Prototype Extraction.}  
In specific spatial scenarios, vessel trajectory modals often exhibit stable sustained intentions due to consistent navigational constraints. Based on this observation, we extract a set of sustained intention prototypes, which are constructed by clustering trajectories from historical scenario $\mathcal{S}^h$, providing prior modals for subsequent tree construction.

To avoid interference between spatial and modal patterns during clustering, we decouple the trajectories into initial points and relative shapes as follows:
\begin{equation}
p_\mathbf{0}=\{(x_1^{\mathcal{T}_1}, y_1^{\mathcal{T}_1}), (x_1^{\mathcal{T}_2}, y_1^{\mathcal{T}_2}),\ldots,(x_1^{\mathcal{T}_\mathrm{m}},y_1^{\mathcal{T}_\mathrm{m}})\}, \quad \mathcal{T}_i \in \mathcal{S}^h,
\end{equation}
\begin{equation}
p_{\mathcal{S}^h - \mathbf{0}} = \mathcal{S}^h - p_\mathbf{0}=\{\{(0,0), (x_2^{\mathcal{T}_i} - x_1^{\mathcal{T}_i}, y_2^{\mathcal{T}_i} - y_1^{\mathcal{T}_i}), \ldots\}\}, 
\end{equation}
where $(x_1^{\mathcal{T}_i},y_1^{\mathcal{T}_i})$ is the first point of $\mathcal{T}_i$, $p_\mathbf{0} \in \mathbb{R}^{(m, 2)}$ and $p_{\mathcal{S}^h - \mathbf{0}} \in \mathbb{R}^{(m, L, 2)}$.

We train an encoder $E_h$ and a decoder $D_h$ to jointly encode $p_\mathbf{0}$ and $p_{\mathcal{S}^h - \mathbf{0}}$ features by reconstructing $\mathcal{S}^h$. Specifically, we construct a latent representation $z^h$ by concatenating $E_h(p_{\mathcal{S}^h - \mathbf{0}})$ with $p_\mathbf{0}$, leading to the reconstruction, as
\begin{equation}
    \hat{\mathcal{S}}^h = D_\mathrm{h} (z^h) = D_\mathrm{h} (E_\mathrm{h}(p_{\mathcal{S}^h - \mathbf{0}}) \oplus p_\mathbf{0}),
\end{equation}
where $\oplus$ is concatenation. Both $p_{\mathcal{S}^h - \mathbf{0}}$ and $p_\mathbf{0}$ are sets encoded via row-wise application of $E_\mathrm{h}$, i.e., each element is encoded independently. This row-wise encoding approach is consistently used throughout the paper for all set encodings.

\begin{algorithm}[!h]  
\caption{Sustained Intention Tree Construction}    
\label{alg:continued_intention_tree}    
\begin{algorithmic}[1]    
\Statex \hspace{-\leftmargin} \textbf{Input:} Historical scenario $\mathcal{S}^h$, Observed scenario $\mathcal{S}^o$, Branch size $k$, Cluster size $\mathcal{C}$
\State \textit{Split $\mathcal{S}^h$ into initial points and relative shapes:}
\State $p_\mathbf{0} = \{(x_1^{\mathcal{T}_1}, y_1^{\mathcal{T}_1}), \ldots, (x_1^{\mathcal{T}_m}, y_1^{\mathcal{T}_m})\}$, where $\mathcal{T}_i \in \mathcal{S}^h$
\State $p_{\mathcal{S}^h - \mathbf{0}} = \mathcal{S}^h - p_\mathbf{0} = \{\{(0, 0), (x_2 - x_1, y_2 - y_1), \ldots\}\}$
\State \textit{Encode each trajectory:} $z^h = E_\mathrm{h}(p_{\mathcal{S}^h - \mathbf{0}}) \oplus p_\mathbf{0}$
\State \textit{Cluster $z^h$ into $\mathcal{C}$ groups and decode centroids:}
\State $\Bar{z}^h \leftarrow$ clustering centroids of $z^h$
\State $\mathcal{P} = D_\mathrm{h}(\Bar{z}^h) = \{p_1, p_2, \ldots, p_\mathcal{C}\}$
\State \textit{Encode observed scenario and prototype set:}
\State $z^o = E_\mathrm{o}(\mathcal{S}^o)$,\quad $z^i = E_\mathrm{i}(\mathcal{P})$
\State \textit{Compute matching probability:} $\Tilde{y} = A_\mathrm{c}(z^o, z^i) \in \mathbb{R}^{(m, \mathcal{C})}$
\State \textit{Select top-$k$ prototype encodings and construct sustained intention tree:}
\State \textit{Construct sustained intention tree:}
\State $\mathcal{B}^k = z^o \oplus \text{top-}k(z^{i}) = \{b_1, b_2, \ldots, b_k\}$
\Statex \hspace{-\leftmargin} \textbf{Output:} Sustained intention tree $\mathcal{B}^k \in \mathbb{R}^{(m, k, 2d)}$
\end{algorithmic}
\end{algorithm}

Then, we cluster the latent feature $z^h$ and decode its centroids $\Bar{z}^h$ to capture representative modals, 
\begin{equation}
    \mathcal{P} = D_\mathrm{h} (\Bar{z}^h) = \{p_1,p_2,\ldots,p_\mathcal{C}\},
\end{equation}
where $\Bar{z}^h = \{ \Bar{z}^{h}_1, ..., \Bar{z}^{h}_{\mathcal{C}}\}$, each $p_i\in\mathbb{R}^{(L, 2)}$ is the $i$-th sustained intention prototype, $\mathcal{P} = \{p_1, p_2, \ldots, p_\mathcal{C}\}$ is the set of sustained intention prototypes, and each $p_i \in \mathbb{R}^{(L, 2)}$ represents a typical modal of length $L$ (longitude-latitude coordinates).

We use the same encoding process and the fitted clustering method to assign each $\mathcal{T}_i$ in $\mathcal{S}$ to its corresponding prototype label $y_i$, and obtain the ground truth label set $\mathcal{Y} = \{y_1, y_2, \ldots, y_m\}$, which serves as the supervision signal for subsequent training.

\textbf{Top-$\mathbf{k}$ Branch Intention Tree Construction.}
The sustained intention tree is an abstract structure constructed at the encoding level by matching each $\mathcal{T}^o$ in observed scenario $\mathcal{S}^o$ with sustained intention prototypes in $\mathcal{P}$, selecting the top-$k$ candidates that represent the most likely future modals.

To model such correlations under multimodality, we adopt a multi-class classification framework based on cross attention $(A_\mathrm{c})$\cite{ref28} to rank and select the most relevant prototypes, i.e.,
\begin{equation}
    \Tilde{y} = A_\mathrm{c} (z^o, z^i) = A_\mathrm{c} (E_\mathrm{o} (S^o), E_\mathrm{i}(\mathcal{P})),
 \end{equation}
where $\Tilde{y} \in \mathbb{R} ^ {(m, C)}$ is the probabilities related to different prototypes, with $z^o \in \mathbb{R}^{(m, d)}$ and $z^i \in \mathbb{R}^{(C, d)}$ serving as query and key, respectively. Here, $m$ represents $m$ trajectories within the scenario, $C$ denotes the total number of sustained intention prototypes in the set $\mathcal{P}$, and $d$ is the hidden dimension of the encoder outputs, which is
shared across all encoder modules in our framework. 
The prototype set $\mathcal{P} \in \mathbb{R}^{(C, L, 2)}$ is encoded row-wise by $E_\mathrm{i}$, producing a key matrix of shape $(C, d)$. Detailed dimensions of all components are summarized in Table~\ref{table_architecture}.

Sequently, we select the top-$k$ rows from $z^i$ according to $\Tilde{y}$, and concatenate the observed scenario $\mathcal{S}^o$ and the top-$k$ rows of the $z^i$ to construct the sustained tree, as
\begin{equation}
\mathcal{B}^k = z^o \oplus \text{top-}k(z^{i}) = \{b_1, b_2, \ldots, b_k\},
\end{equation}
where $\mathcal{B}^k \in \mathbb{R}^{(m, k, 2d)}$ is the encoded sustained intention tree, and each branch $b_i \in \mathbb{R}^{(m, 2d)}$ corresponds to a potential future path.

\subsection{Global Transient Intention Optimization}
\label{subsec:global_transient_intention}
As the final decoding stage, this module performs global trajectory prediction based on the preceding output. It consists of two components: (1) destination-aware interaction CVAE, and (2) global intention-guided trajectory generation.(Algorithm~\ref{alg:global_instant_intention})

\textbf{Destination-based Interaction CVAE. }
To capture the interaction between sustained and transient intentions, we adopt a destination prediction modeling strategy. The predicted destination serves as a compact representation of their integrated effect over the entire trajectory.

Specifically, given the destination $\mathcal{D} \in \mathbb{R}^{(m, 2)}$, which is the final point of the trajectories in scenario, where $2$ corresponds to the latitude and longitude coordinates, we construct a latent feature $z^\mathrm{lat}$, i.e.,
\begin{equation}
\label{encode_latent}
    z^\mathrm{lat}  = E_\mathrm{l} (E_\mathrm{d} (\mathcal{D}) \oplus \mathcal{B}^k),
\end{equation}
where $z^\mathrm{lat} \in \mathbb{R}^{(m, k, 2d)}$ denotes the latent variable from the CVAE, and split it into two parts: $\mu$ and $\sigma$ which are learnable features representing the mean and standard deviation for each of the $k$ branches, as,
\begin{equation}
    \mu = z^\mathrm{lat}_{*, :d}, \quad \sigma = z^\mathrm{lat}_{*, d:},
\end{equation}
where $\mu \in \mathbb{R}^{(m, k, d)}$ and $\sigma \in \mathbb{R}^{(m, k, d)}$.

\begin{algorithm}[h]
\caption{Global Transient Intention Optimization}    
\label{alg:global_instant_intention}    
\begin{algorithmic}[1]    
\Statex \hspace{-\leftmargin} \textbf{Input:} \textit{Sustained intention tree} $\mathcal{B}^k$, \textit{Ground truth destination} $\mathcal{D}$, \textit{Encounter mask} $M$
\State \textit{Encode destination and compute latent distribution:}
\State \hspace{1em} $z^\mathrm{lat} = E_\mathrm{l}(E_\mathrm{d}(\mathcal{D}) \oplus \mathcal{B}^k)$
\State \hspace{1em} $\mu =z^\mathrm{lat}_{*, :d} \quad \sigma = z^\mathrm{lat}_{*, d:}$
\State \textit{Optimize the $\mathcal{N}(\mu, \sigma)$ to $\mathcal{N}(0, \epsilon \cdot I)$}
\State \textit{Sample latent variables:} $z \sim \mathcal{N}(0, \epsilon \cdot \mathrm{I})$
\State \textit{Decode destination predictions:} $\hat{\mathcal{D}} = D_\mathrm{d}(z \oplus \mathcal{B}^k)$
\State \textit{Fuse with sustained intention tree:} 
\State \hspace{1em} $z^{\prime} = \mathcal{B}^k \oplus E_\mathrm{d}(\hat{\mathcal{D}})$
\For {each attention layer $l = 1$ to $L_d$}
    \State $z^{\prime} = z^{\prime} + A_{\mathrm{n}, l}(z^{\prime}, M) \quad l = 1, 2, \ldots, L_d$
\EndFor
\State \textit{Generate final multimodal predictions:} $S^p = D_\mathrm{p}(z^{\prime})$
\Statex \hspace{-\leftmargin} \textbf{Output:} Predicted trajectories $S^p$, predicted destinations $\hat{\mathcal{D}}$
\end{algorithmic}    
\end{algorithm}

Due to the unavailability of the true destination, we optimized the $\mathcal{N}(\mu, \sigma)$ to approximate the standard normal distribution $\mathcal{N}(0, \epsilon \cdot \mathrm{I})$. During prediction, we directly sample $n$ latent variables from the standard normal distribution $z \sim \mathcal{N}(0, \epsilon \cdot \mathrm{I})$\cite{ref26}. Specifically, for $m$ trajectories and $k$ sustained intention branches, we perform $\frac{n}{k}$ samplings per branch. Here, $z \in \mathbb{R}^{(m, k, \frac{n}{k}, 2d)}$ is the sampled latent representation, $\mathrm{I} \in \mathbb{R}^{(2d, 2d)}$ is the identity matrix, and $\epsilon$ is a positive scalar (e.g., 1.3) that controls the sampling variance.

Subsequently, the sampled \( z \) is concatenate with the sustained intention tree as in Eq.~\eqref{encode_latent} and fed into the decoder to predict the destination, jointly leveraging both features as follows:
\begin{equation}
\hat{\mathcal{D}} = D_\mathrm{d} \left(z \oplus \mathcal{B}^k\right),
\end{equation}
where $\hat{\mathcal{D}} \in \mathbb{R}^{\left(m, k, \frac{n}{k}, 2\right)}$ is the set of predicted destinations.

\textbf{Global Intention Prediction. }
Given the encounter mask $M \in \mathbb{R}^{m \times m}$, the sustained intention tree $\mathcal{B}^k$ and the predicted destination $\hat{\mathcal{D}}$, we construct a fused latent feature 
\begin{equation}
    z^{\prime}=\mathcal{B}^k \oplus E_\mathrm{d}(\hat{\mathcal{D}}),
\end{equation}
for multimodal trajectory prediction, where $z^{\prime} \in \mathbb{R}^{(m, k, \frac{n}{k}, 3d)}$ with $m$ for the number of trajectories in scenario, $k$ for branches, $\frac{n}{k}$ for sampled destinations per branch, and $d$ as the embedding dimension.

However, conflicts may arise when integrating multiple sustained intention branches, potentially leading to inconsistent global predictions. To address this, we adopt a non-local attention mechanism that improves coherent and globally consistent trajectory generation.

The non-local attention mechanism is applied iteratively with a depth of $L_d$, progressively integrating contextual information across all branches. At each layer, attention weights modulated by $M$ are added to $z^{\prime}$, emphasizing multi-scale correlations. Specifically, 
\begin{equation}
z^{\prime} = z^{\prime} + A_{\mathrm{n}, l}(z^{\prime}, M), \quad l = 1, 2, \ldots, L_d,
\end{equation}
where $M$ guides attention computation by restricting it to meaningful vessel pairs.

By learning the global relationships between $z^{\prime}$ which are highlighted by $M$, DI-MTP can effectively capture the characteristics of encounters and guide the prediction towards a safe trajectory. We validate the model's ability to handle encounters in Fig.~\ref{fig_encounter_pred}.

Finally, the multimodal predictions $\mathcal{T}^p$ is generated by
\begin{equation}
    S^p = D_\mathrm{p} (z^{\prime}), \quad S^p \in \mathbb{R}^{(m, k, \frac{n}{k}, L, 2)},
\end{equation}
where $m$ representing a single observed trajectory, 
$k$ the number of sustained intention branches, 
$\frac{n}{k}$ the number of sampled trajectories per branch, 
$L$ the number of predicted time steps, 
and $2$ corresponding to the longitude and latitude at each step.

\subsection{Loss Function}
Our training strategy consists of two stages. First, we train the autoencoder to extract the sustained intention prototypes by reconstructing each $\mathcal{T}^h$, using the reconstruction loss:
\begin{equation}
\mathcal{L}_\mathrm{rec} = (S^h-\hat{S}^h)^2,
\end{equation}
where $S^h$ and $\hat{S}^h$ represent the original and reconstructed history scenarios.

For the DI-MTP modules, we incorporate classification, destination prediction, and final prediction losses.
For selecting the top $k$ branches in constructing the sustained intention tree, we use a classification loss $\mathcal{L}_\mathrm{clf}$ based on cross-entropy, which helps the model choose the highest relevant sustained intentions for further prediction. The $\mathcal{L}_\mathrm{clf}$ is defined as:
\begin{equation}
\mathcal{L}_\mathrm{clf} = \mathrm{CE}(\mathcal{Y}, \hat{\mathcal{Y}}),
\end{equation}
where $\mathrm{CE}$ denotes the cross-entropy loss, $\hat{\mathcal{Y}} = \{\hat{y_1}, \hat{y_2}, \ldots, \hat{y_m}\}$ represents the set of clustered class labels for each $\mathcal{T}$ in the set $\mathcal{S}$, and $\mathcal{Y}$ is the corresponding ground truth class label mentioned in Section \ref{subsec:sustained_intention_tree}.

The destination-based interaction CVAE, whose loss consists of two components: the KL divergence and the destination prediction error, defined as:
\begin{equation}
\mathcal{L}_\mathrm{cvae} = \mathrm{KL}(\mathcal{N}(\mu, \sigma) \| \mathcal{N}(0, \epsilon \cdot \mathbf{I})) + \alpha \cdot\|\mathcal{D} - \hat{\mathcal{D}}\|^2,
\end{equation}
where $\mathrm{KL}$ represents the Kullback-Leibler divergence between the predicted latent distribution $\mathcal{N}(\mu, \sigma)$ and the prior $\mathcal{N}(0, \epsilon \cdot \mathbf{I})$. The second term measures the reconstruction error of the predicted destination $\hat{\mathcal{D}}$ against the ground truth $\mathcal{D}$ and the coefficient $\alpha$ is a balancing hyperparameter.

The final module is the global intention prediction, where we adopt a regression loss $\mathcal{L}_\mathrm{reg}$ to minimize the error between the predicted and ground truth trajectories. During training, we use the ground-truth class label $\mathcal{Y}$ to select the closest sustained prototype and sample once from the latent space ($n = k = 1$), generating a single trajectory per instance. The loss is defined as:
\begin{equation}
\mathcal{L}_\mathrm{reg} = \sum_{i=1}^{m} \|\mathcal{T}^p_i - \hat{\mathcal{T}^p_i}\|^2,
\end{equation}
where $\hat{\mathcal{T}^p_i} \in \mathbb{R}^{(n, L_p, 2)}(n=1)$ and $\mathcal{T}^p_i \in \mathbb{R}^{(L_p, 2)}$ represent the predicted and ground truth trajectories.

For DI-MTP training, we combine the losses for different modules to an overall loss, which is a weighted sum of the individual losses:
\begin{equation}
\mathcal{L}_\mathrm{DI-MTP} = \lambda_1 \mathcal{L}_\mathrm{clf} + \lambda_2 \mathcal{L}_\mathrm{cvae} + \lambda_3 \mathcal{L}_\mathrm{reg},
\end{equation}
where $\lambda$s are the balancing parameters, controlling the importance of each modules.

\textbf{Computational Complexity.}
The overall computational complexity of DI-MTP consists of: (1) an offline prototype extraction step with complexity $\mathcal{O}(dmCI)$, and (2) an online inference process with complexity $\mathcal{O}(dmk + dm^2)$. Here, $m$ is the number of trajectories, $d$ the feature dimension, $C$ the number of clusters, $k$ the number of selected prototypes, and $I$ the number of clustering iterations.

\textit{1) Prototype Extraction.}
This is a one-time offline process that clusters $m$ historical trajectory features into $C$ prototypes, with complexity $\mathcal{O}(dmCI)$. The resulting prototypes are reused during inference and do not affect runtime performance.
\textit{2) Online Inference.}
At inference time, the model performs cross-attention between $m$ input trajectories and $k$ prototypes with complexity $\mathcal{O}(dmk)$, and pairwise interaction among $m$ trajectories for global intention optimization with complexity $\mathcal{O}(dm^2)$. In practice, both $m$ and $k$ are relatively small (mostly $m \leq 20$, $k \leq 50$).

\section{Experiments}  
  
\subsection{Dataset}
We use real AIS data from two distinct maritime regions to ensure a comprehensive evaluation of our model across diverse navigation scenarios.

\subsubsection{Zhoushan}
The Port of Ningbo-Zhoushan features diverse maritime traffic with various vessel types—commercial ships, cargo vessels, tankers, fishing boats—and complex interaction scenarios involving single and multiple encounters. AIS data, resampled at 30-second intervals, spans January to March 2022, including 331 training and 142 testing scenarios.

\subsubsection{Hainan}
Hainan coastal waters have intricate navigation routes and dense traffic, including ferries, cargo ships, and cruise liners. The dataset captures challenging scenarios such as dense vessel encounters near ports and narrow waterway crossings. Similarly, the AIS data is resampled at 30 seconds, covers October 2018.
  
\subsection{Implementation Details}  
\label{config}
We set the size of sustained intention prototypes $\mathcal{C} = 30$, the size of selected highly-related intentions $k = 10$, and the number of predicted trajectories per observed trajectory $n=50$, respectively. All modules use MLPs with ReLU activation. Table \ref{table_architecture} details the network architecture, where the overall trajectory length $L=18$, observed length $L_o=6$, prediction length $L_p=12$, and hidden dimension $d=64$. For DI-MTP training, we use the Adam optimizer with a learning rate of \(1 \times 10^{-4}\) and a multi-step scheduler. The learning rate is multiplied by 0.5 at predefined epochs (160 and 300) over a total of 500 epochs. The coefficient $\alpha$ in $\mathcal{L}_\mathrm{cvae}$ is 0.01. And the coefficients in $\mathcal{L}_\mathrm{DI-MTP}$ are set to \(\lambda_1=\lambda_2=\lambda_3=1\), assigning equal weights to the three objectives, which are designed to jointly enable accurate and interpretable trajectory forecasting. This balanced setup offers a consistent and robust baseline for evaluating the proposed unified framework.

\begin{table}[!t]
\renewcommand{\arraystretch}{1.6}
\caption{Detailed Network Architecture Configuration}
\label{table_architecture}
\centering
\setlength{\aboverulesep}{0.4ex}
\setlength{\belowrulesep}{0.4ex}
\begin{tabular}{@{}m{0.19\columnwidth}m{0.28\columnwidth}>{\raggedright\arraybackslash}m{0.43\columnwidth}@{}}
\toprule
\textbf{Module} & \textbf{Description} & \textbf{Layer Configuration} \\
\hline
$E_\mathrm{h}$ & History trajectory encoder & \makecell[l]{$L \times 2 \rightarrow$ 1024 $\rightarrow$ 512 $\rightarrow$\\1024 $\rightarrow$ $d$} \\
\hdashline[1pt/2pt]
$D_\mathrm{h}$ & History trajectory decoder & \makecell[l]{$d + 2$ $\rightarrow$ 1024 $\rightarrow$ 512 $\rightarrow$\\1024 $\rightarrow L \times 2$} \\
\hdashline[1pt/2pt]
$E_\mathrm{o}$ & Observed trajectory encoder & \makecell[l]{$L_o \times 2 \rightarrow$ 1024 $\rightarrow$ 512 $\rightarrow$\\1024 $\rightarrow$ $d$} \\
\hdashline[1pt/2pt]
$E_\mathrm{i}$ & Intention prototype encoder & \makecell[l]{$L \times 2 \rightarrow$ 1024 $\rightarrow$ 512 $\rightarrow$\\1024 $\rightarrow$ $d$} \\
\hdashline[1pt/2pt]
$E_\mathrm{d}$ & Destination encoder & \makecell[l]{2 $\rightarrow$ 1024 $\rightarrow$ 512 $\rightarrow$\\1024 $\rightarrow$ $d$} \\
\hdashline[1pt/2pt]
$D_\mathrm{d}$ & Destination decoder & \makecell[l]{$d \times 4$ $\rightarrow$ 1024 $\rightarrow$ 512 $\rightarrow$\\1024 $\rightarrow$ 2} \\
\hdashline[1pt/2pt]
$E_\mathrm{l}$ & Latent encoder & \makecell[l]{$d \times 3 \rightarrow$ 1024 $\rightarrow$ 512 $\rightarrow$\\1024 $\rightarrow d \times 2$} \\
\hdashline[1pt/2pt]
\makecell[l]{$A_\mathrm{n}$} & Non-local attention & \makecell[l]{Q: $d \times 3 \rightarrow$ 1024 $\rightarrow$ 512 $\rightarrow$\\1024 $\rightarrow$ 1024\\[1ex]K: $d \times 3 \rightarrow$ 1024 $\rightarrow$ 512 $\rightarrow$\\1024 $\rightarrow$ 1024\\[1ex]V: $d \times 3 \rightarrow$ 1024 $\rightarrow$ 512 $\rightarrow$\\1024 $\rightarrow d \times 3$} \\
\hdashline[1pt/2pt]
\makecell[l]{$A_\mathrm{c}$} & Cross attention & \makecell[l]{Q: $d$ $\rightarrow$ 1024 $\rightarrow$ 1024 $\rightarrow$ $d$\\[1ex]K: $d$ $\rightarrow$ 1024 $\rightarrow$ 1024 $\rightarrow$ $d$\\[1ex]} \\
\hdashline[1pt/2pt]
$D_\mathrm{p}$ & Predictor & \makecell[l]{$d \times 3 \rightarrow$ 1024 $\rightarrow$ 512 $\rightarrow$\\1024 $\rightarrow L_p \times 2$} \\
\hline
\end{tabular}
\end{table}

\subsection{Evaluation Metrics}  
For prediction evaluation, we use the Average Displacement Error (ADE) and Final Displacement Error (FDE) metrics \cite{ref7,ref23,ref24,ref25,ref26}. ADE represents the average distance between predicted trajectories and ground truths, while FDE measures the distance between predicted destinations and real destinations. For coordinates $x$ and $y$ representing longitude and latitude respectively, we calculate the distance in meters as:
\begin{equation}
\begin{split}
d((x^{\mathcal{T}_i}_t, y^{\mathcal{T}_i}_t), (x^{\mathcal{T}_j}_t, y^{\mathcal{T}_j}_t))=2r \cdot \arcsin\Bigg(\sqrt{\sin^2\left(\frac{y^{\mathcal{T}_j}_t-y^{\mathcal{T}_i}_t}{2}\right)} \\
+ \cos(y^{\mathcal{T}_i}_t) \cdot \cos(y^{\mathcal{T}_j}_t) \cdot \sin^2\left(\frac{x^{\mathcal{T}_j}_t-x^{\mathcal{T}_i}_t}{2}\right) \Bigg),
\end{split}
\end{equation}
where \(r = 6,371,000\) meters (Earth's radius), and $(x^{\mathcal{T}_i}_t, y^{\mathcal{T}_{i}}_t)$ and $(x^{\mathcal{T}_j}_t, y^{\mathcal{T}_{j}}_t)$ are the vessel's position in the $i$-th and $j$-th trajectory at $t$ step, $i, j \in [1, .., \mathrm{m}]$. The ADE and FDE are calculated as:
\begin{equation}
\text{ADE} = \sum_{i=1}^{m} \min_{1 \leq j \leq n} \frac{1}{L_p} \sum_{t = L_o+1}^{L_o + L_p} d\left((x^{\hat{\mathcal{T}^p_{i,j}}}_t, y^{\hat{\mathcal{T}^p_{i,j}}}_t), (x^{\mathcal{T}^p_i}_t, y^{\mathcal{T}^p_i}_t)\right),
\end{equation}

\begin{equation}
\text{FDE} = \sum_{i=1}^{m} \min_{1 \leq j \leq n} d\left( (x^{\hat{\mathcal{T}^p_{i,j}}}_{L_o + L_p},\ y^{\hat{\mathcal{T}^p_{i,j}}}_{L_o + L_p}),\ (x^{\mathcal{T}^p_i}_{L_o + L_p},\ y^{\mathcal{T}^p_i}_{L_o + L_p}) \right),
\end{equation}
where $L_p$ is the prediction length, $\mathcal{T}^p_i \in \mathbb{R}^{(L_p, 2)}$ denotes the $i$-th ground truth trajectory, and $\mathcal{T}^p_{i,j} \in \mathbb{R}^{(L_p, 2)}$ is its $j$-th predicted candidate, here $1 \leq i \leq m$ and $1 \leq j \leq n$. In practice, both ADE and FDE are computed using the same index $j$ that minimizes the combined error $\text{ADE} + \text{FDE}$.

\subsection{Results and Analysis}  
  
\subsubsection{Baselines}  
To evaluate the performance and robustness of DI-MTP, we use the following baseline models:

For DTP methods, we include vessel trajectory prediction models, Zhang et al.~\cite{ref30} and Capobianco et al.~\cite{ref31}. We also include TBENet~\cite{ref32}, a bi-directional information fusion-driven prediction model which represents a state-of-the-art (SOTA) approach specifically tailored for vessel trajectory prediction in intelligent maritime transportation systems.

For MTP methods, to the best of our knowledge, only GRU-CVAE~\cite{ref7} and VT-MDM~\cite{ref8} have been specifically developed for the maritime domain. However, both methods are limited in applicability, as the former focuses exclusively on two-vessel encounters, while the latter does not explicitly consider encounters at all. Moreover, neither of these methods is publicly available. Therefore, we reproduce two versions of GRU-CVAE~\cite{ref7} for evaluation: the GRU-CVAE-base, directly corresponding to its seq2seq variant, and the GRU-CVAE, built upon the full model proposed in the original paper, with a modified vessel history encoder to support multi-vessel scenarios. To better benchmark and evaluate the effectiveness of MTP methods, we further include PECNet~\cite{ref26}, SIT~\cite{ref24}, and TUTR~\cite{ref25}.

\subsubsection{Quantitative Evaluation}
Table \ref{table_model_comparison} presents performance comparisons across the Zhoushan and Hainan maritime datasets.
\paragraph{Comparison with DTP Methods}
MTP methods substantially outperform deterministic models Zhang et al., Capobianco et al., and TBENet on both datasets. DI-MTP significantly outperforms TBENet, reducing ADE/FDE by 20.3\%/25.2\% on Zhoushan and 45.2\%/43.1\% on Hainan. The significant gaps in ADE and FDE indicate that conventional approaches are inadequate in capturing the multimodal nature of complex maritime scenarios, which play a vital role in ensuring maritime safety.
\paragraph{Comparison with MTP Methods}
DI-MTP also consistently surpasses other MTP baselines. Compared with the TUTR, DI-MTP achieves ADE/FDE reductions of 18.28\%/8.51\% on Zhoushan and 27.48\%/27.40\% on Hainan. In contrast, GRU-CVAE, though capable of modeling stochasticity via latent variables, often collapses to limited modals, leading to higher prediction errors~(further analyzed in Fig.~\ref{fig_multimodality}). Specifically, compared to GRU-CVAE, DI-MTP achieves ADE/FDE reductions from 83.11/129.98 to 64.91/102.33 on Zhoushan (21.89\%/21.27\%), and from 65.37/94.33 to 42.87/69.07 on Hainan (34.42\%/26.77\%). The results demonstrate the strength of our dual-intention modeling in capturing both sustained and transient behaviors. (See Fig.~\ref{fig_combined_trends} for details of sustained and transient intention analysis).

\setlength{\tabcolsep}{2.5pt}        
\setlength{\aboverulesep}{0.5ex}   
\setlength{\belowrulesep}{0.3ex}   
\setlength{\abovetopsep}{0pt}      

\begin{table*}[!t]
\centering
\caption{Performance Comparison of Trajectory Prediction Models (ADE/FDE; lower is better for both metrics)}
\label{table_model_comparison}
{\renewcommand{\arraystretch}{1.5} 
\begin{tabular}{cccccccccc}
\toprule
\diagbox{Dataset}{Model} & \makecell{Zhang\\et al.~\cite{ref30}} 
                         & \makecell{Capobianco\\et al.~\cite{ref31}} 
                         & \makecell{TBENet\\\cite{ref32}} 
                         & \makecell{PECNet\\\cite{ref26}} 
                         & \makecell{SIT\\\cite{ref24}} 
                         & \makecell{TUTR\\\cite{ref25}} 
                         & \makecell{GRU-CVAE\\-base~\cite{ref7}} 
                         & \makecell{GRU-CVAE\\~\cite{ref7}} 
                         & \makecell{DI-MTP\\(Ours)} \\
\hline
Zhoushan & 131.96/234.68 & 142.04/246.66 & 81.50/136.83   & 118.33/245.70 & 83.57/129.54 & 79.43/111.85 & 138.41/129.55 & 83.11/129.98 & \textbf{64.91/102.33} \\
Hainan   & 108.34/192.29 & 129.84/205.83 & 78.16/121.32   & 74.42/145.57  & 63.97/106.90 & 59.11/95.13  & 146.39/104.79 & 65.37/94.33 & \textbf{42.87/69.07} \\
\hline
\end{tabular}
} 
\vspace{0.5ex} 
\end{table*}

\subsubsection{Impact of Observation and Prediction Horizon}

\begin{figure*}[!t]
\centering
\includegraphics[width=0.45\textwidth]{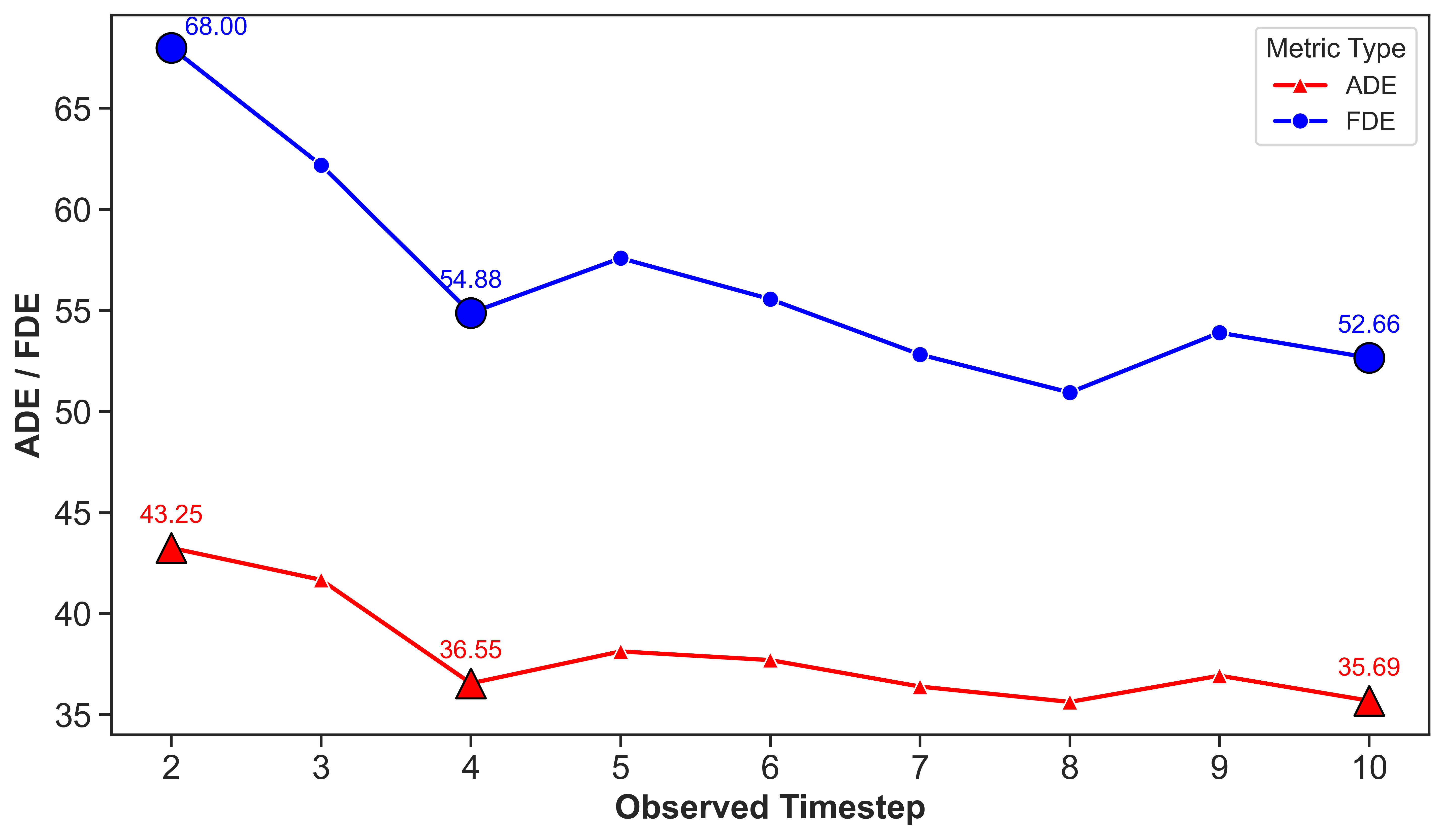} \label{obs_vary}
\hfill
\includegraphics[width=0.45\textwidth]{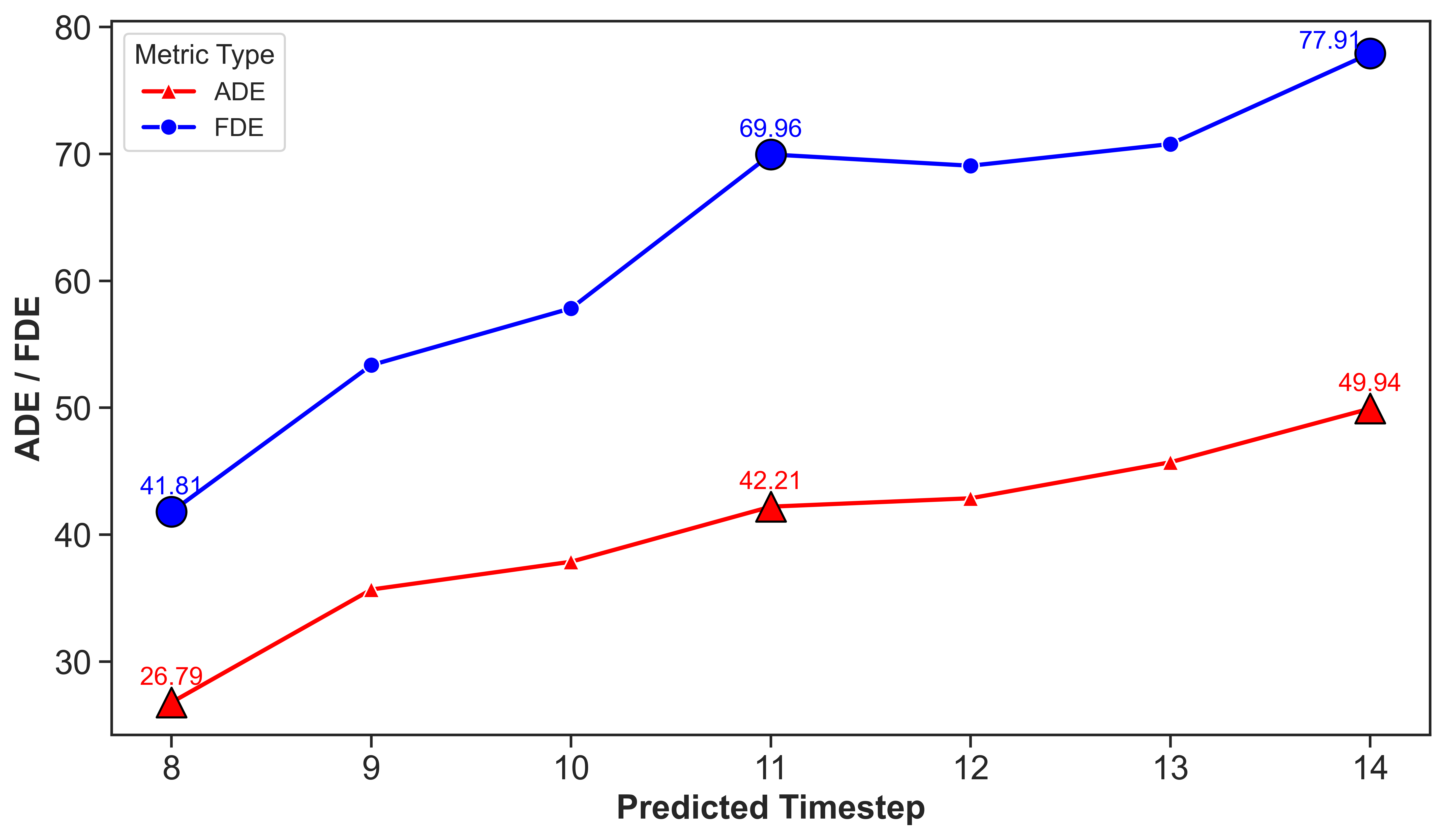} \label{pred_vary}
\caption{Trend curves of ADE and FDE under varying sequence settings: (a) observation lengths from 2 to 10 timesteps (prediction length fixed at 10); (b) prediction lengths from 8 to 14 timesteps (observation length fixed at 6). Lower values indicate better performance.}
\vspace{-3mm}
\label{fig_obs_pred_vary}
\end{figure*}

To assess the temporal robustness of DI-MTP, we examine how varying the observation and prediction lengths affects model performance on the Hainan dataset, while keeping all other settings consistent with Section~\ref{config}. As shown in Fig.~\ref{fig_obs_pred_vary}, we conduct two sets of experiments: (a) fixing the prediction length at 10 timesteps and varying the observation length from 2 to 10; and (b) fixing the observation length at 6 timesteps while varying the prediction length from 8 to 14.

The results demonstrate that DI-MTP maintains robust performance across different temporal configurations. In Fig.~\ref{fig_obs_pred_vary}(b), as the prediction length increases, both ADE and FDE gradually deteriorate—an expected trend given the increasing uncertainty of long-horizon predictions.

In Fig.~\ref{fig_obs_pred_vary}(a), performance improves as observation length increases, due to richer context being available. However, the marginal gain diminishes with longer observations, and fluctuations at certain lengths (e.g., 5 and 9) suggest sensitivity in the model's structure. We attribute this to two factors: (1) the model's sustained intention matching can be effective even with relatively short observation windows, and (2) the fixed number of intention prototypes may limit the model’s capacity to encode increasingly dense information as observation length grows. This reveals the limitation in fixed number prototype-based modeling, indicating a potential improving direction.

\subsubsection{Sustained intention Analysis}  
\label{sustained intention analysis}
To evaluate the effectiveness of sustained intentions in modeling multimodality, we conduct a controlled experiment by setting the number of samples $n$ equal to the number of sustained intention branches $k$—both defined in Section~\ref{Definition3}—thereby isolating the impact of sustained intention while avoiding additional stochasticity from transient variations.

Fig.~\ref{fig_combined_trends}(a) illustrates the relationship between evaluation metrics and the number of considered sustained intentions \(k\). Both ADE and FDE values show a consistent decreasing trend before converging. Recall the ade and fde in Table~\ref{table_model_comparison}, with \(k=1\), DI-MTP still outperforms PECNet in both ADE and FDE. With increasing 
$k$
, DI-MTP consistently improves in performance and surpasses all baselines when 
$k$=10. At this point, it outperforms GRU-CVAE by 16.4\% in ADE and 12.27\% in FDE. This demonstrates that sustained intentions contain rich multimodal information that provides diverse basis modalities for predictions.

These results suggest that sustained intentions effectively handle modal uncertainty, offering rich multimodal cues and supporting diverse predictions. Moreover, sustained intentions consistently outperform baselines across various scenarios, demonstrating strong generalizability and contextual adaptability.

\subsubsection{Global Transient Intention Analysis} 
\label{transient intention analysis}
To assess the effectiveness of transient intentions, we conduct an experiment where only a single sustained intention branch ($k=1$) is used. Specifically, we directly assign the label obtained from clustering as the sole sustained intention prototype, allowing the model to focus exclusively on transient variations during prediction.

\begin{figure*}[!t]
\centering
\includegraphics[width=0.45\textwidth]{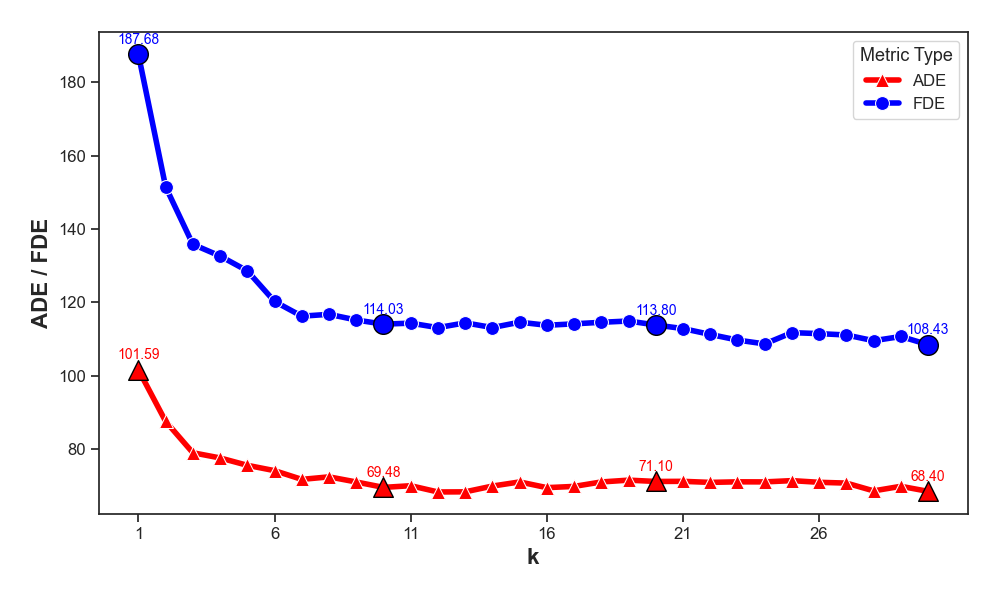} \label{fig_continuous_intention}
\hfill
\includegraphics[width=0.45\textwidth]{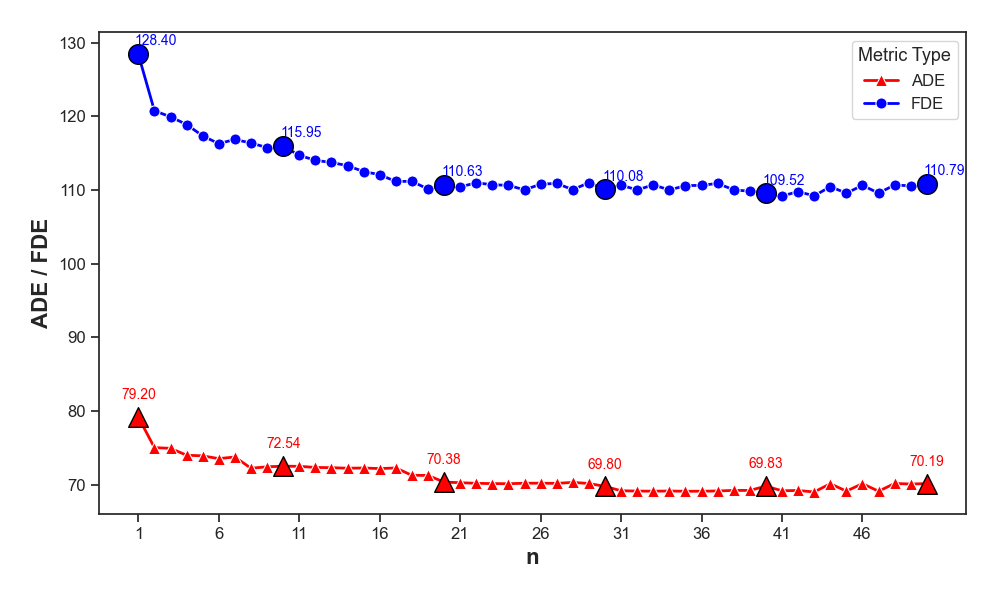} \label{fig_global_instantaneous_intention}
\vspace{-0.5cm}
\caption{Trend curves of ADE/FDE for varying numbers of sustained intentions \(k\) in (a) and varying sampling numbers \(n\) in (b). Lower metric values are better.}
\label{fig_combined_trends}
\end{figure*}

Fig.~\ref{fig_combined_trends}(b) shows the performance trend as the CVAE samples 
$n$ increases, both ADE and FDE metrics decrease. Specifically, at \(n=1\), DI-MTP already outperforms GRU-CVAE with a relative improvement of 4.7\% in ade and 1.22\% in fde. And at \(n=10\), it surpasses all baseline methods.

These results demonstrate the robustness of the destination-based interaction CVAE in modeling the dynamics and uncertainty of interactions between transient and sustained intentions.

\subsubsection{Intention Visualization and Analysis}
\begin{figure*}[!t]  
    \centering
    \includegraphics[width=0.30\textwidth]{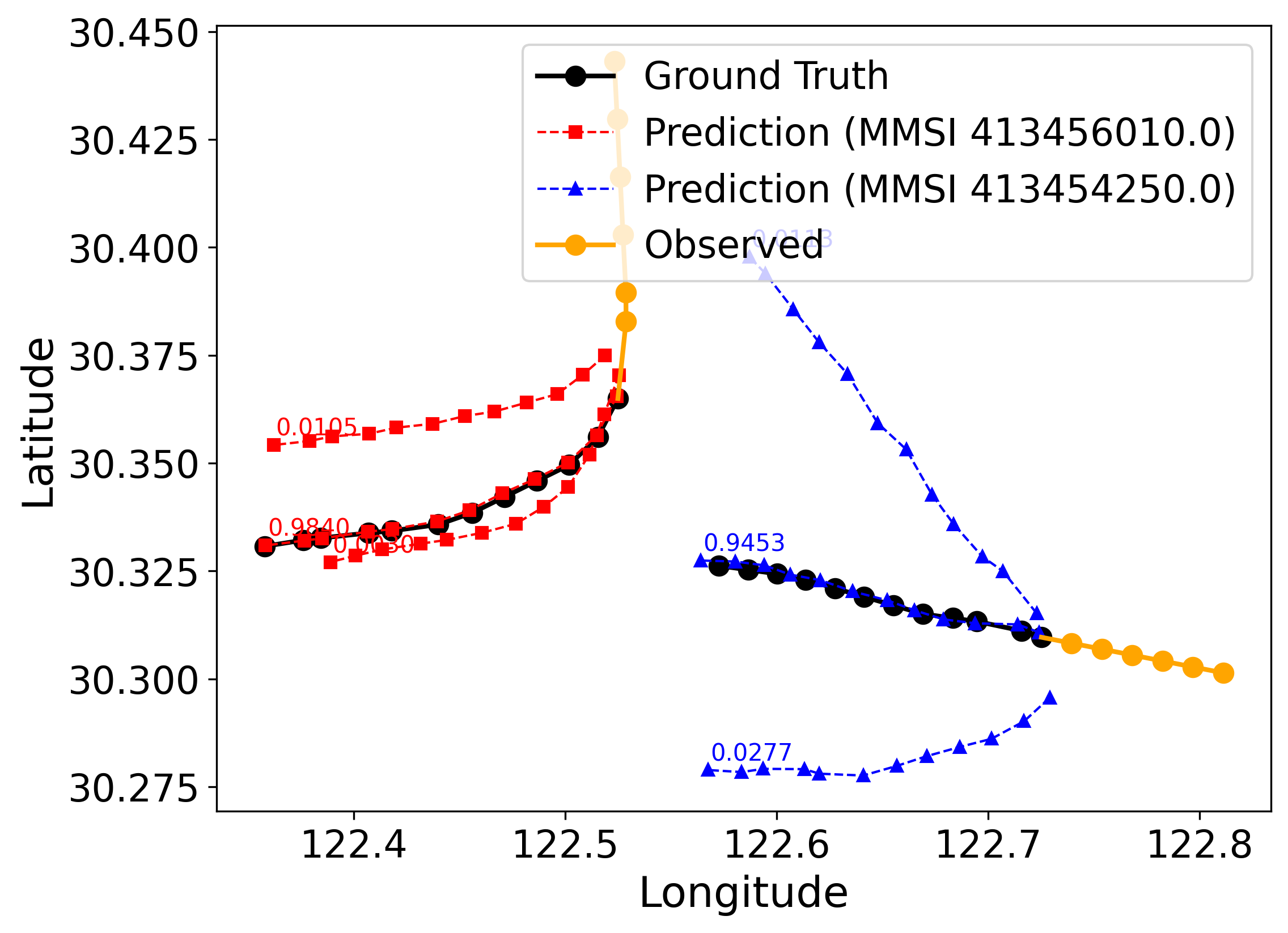}
    \hfill
    \includegraphics[width=0.308\textwidth]{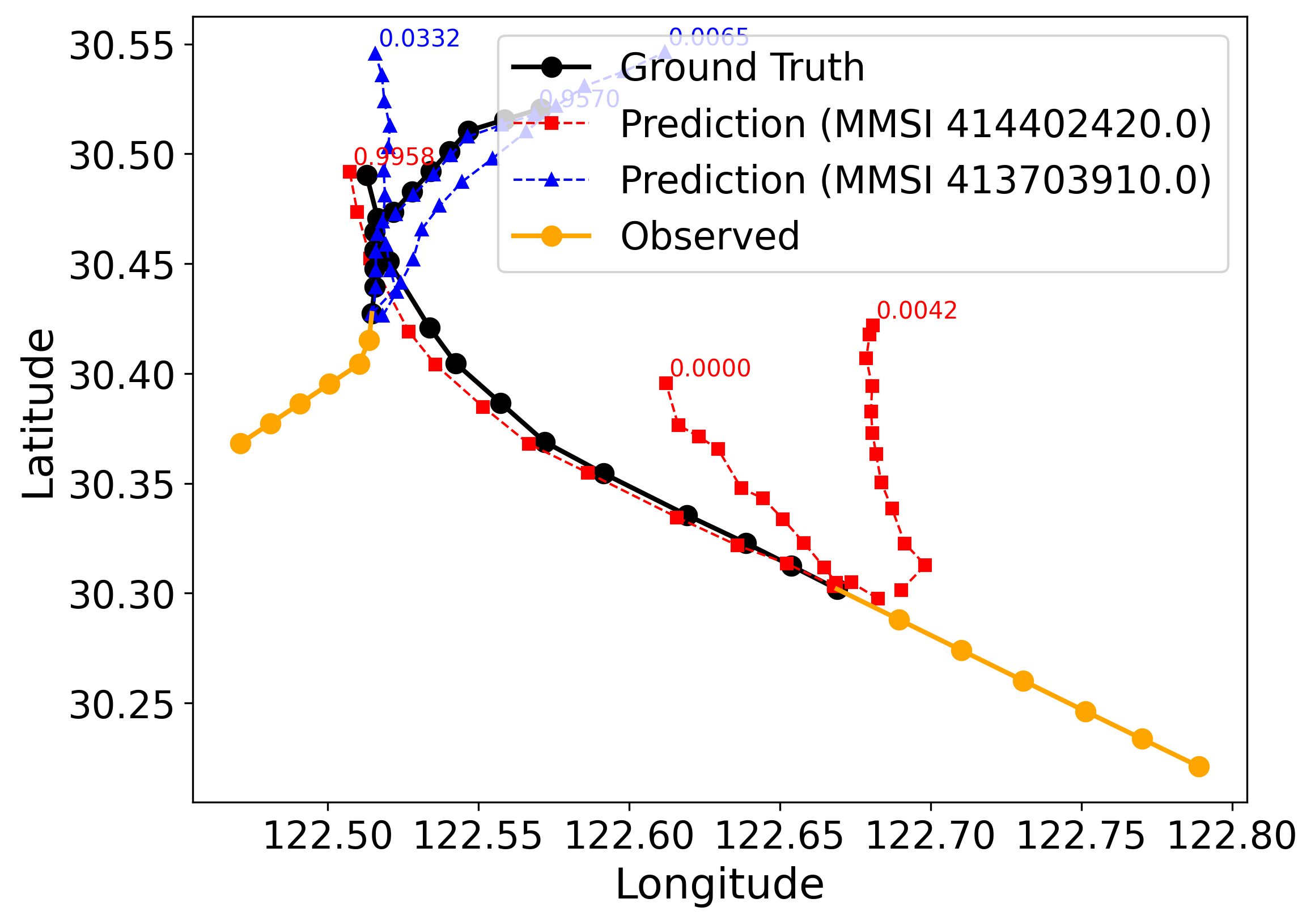}
    \hfill
    \includegraphics[width=0.308\textwidth]{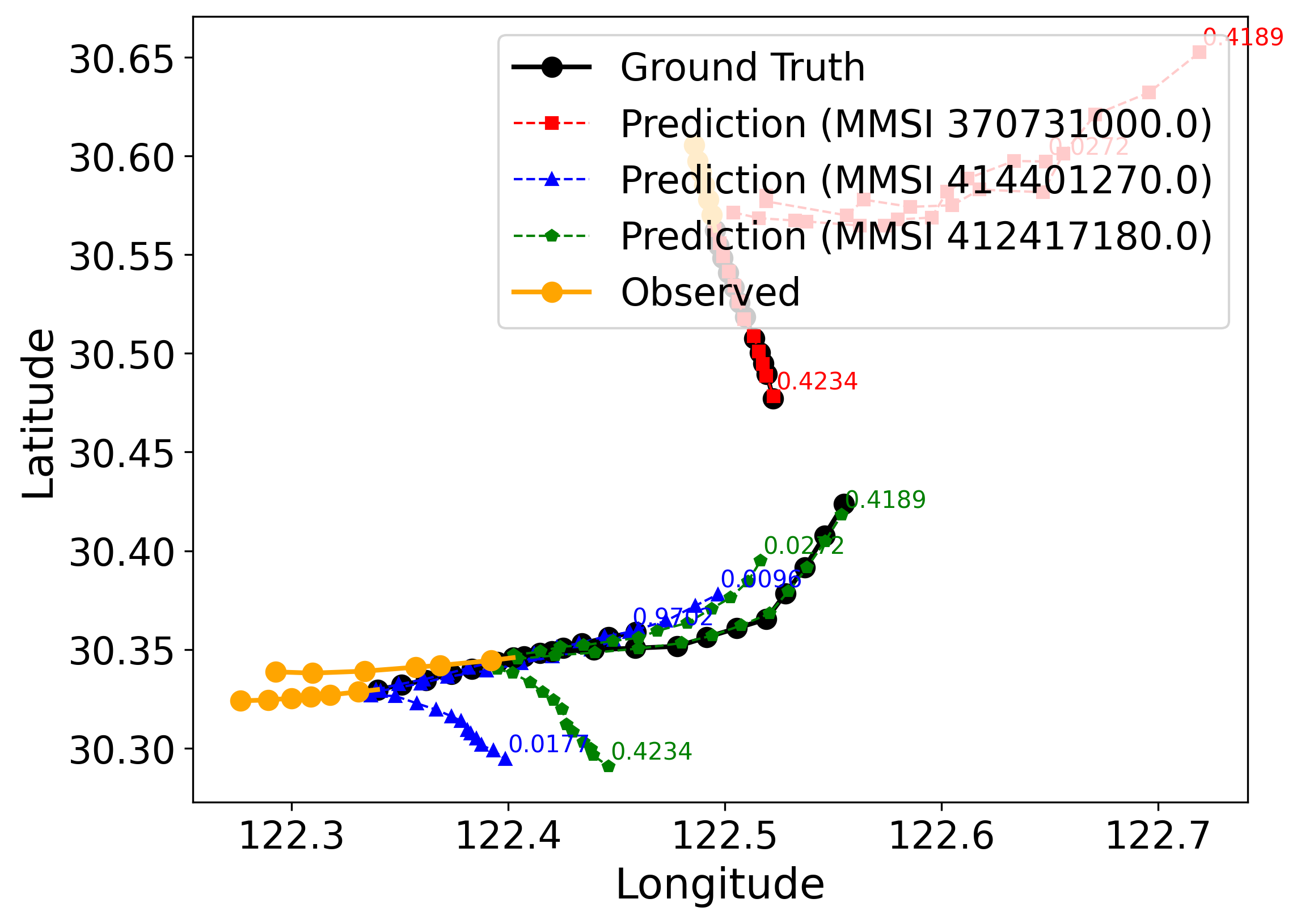}
    \caption{
     Sustained intentions in diverse scenarios, where yellow lines indicate observed trajectories, black lines represent ground truth, and red, green, and blue dashed lines show the top 3 sustained intentions for each vessel with their MMSI in legend, respectively. (a) Non-encounter scenario. (b) Two vessels encounter scenario. (c) Three vessels encounter scenario.
    }
    \label{fig_sustained}
\vspace{-3mm}
\end{figure*}

\label{encounter val}
To better understand how DI-MTP captures and distinguishes between multiple future modalities, we visualize the prediction results under a fixed configuration of $k = n = 3$ across three representative scenarios: non-encounter, two-vessel encounter, and three-vessel encounter, as shown in Fig.~\ref{fig_sustained}.

In the non-encounter and two-vessel encounter cases (Fig.~\ref{fig_sustained}(a) and (b)), the model assigns high confidence to the correct modality, with clearly dominant probabilities, indicating successful sustained intention matching and confident predictions. Notably, even under high-probability conditions, the model maintains multimodal diversity rather than collapsing into a single modal.

\begin{figure*}[!t]  
    \centering
    \subfloat[Two-vessel encounter scenario]{
        \includegraphics[width=\textwidth]{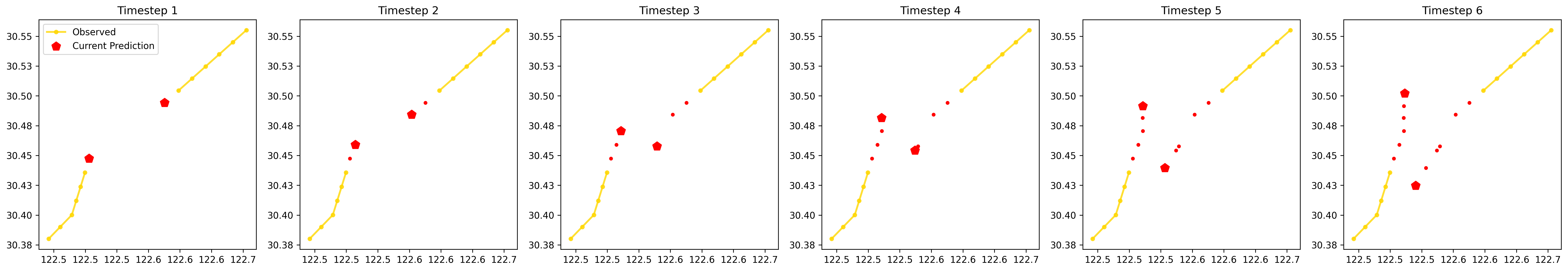}
        \label{fig:two_vessel_timesteps}
    }
    \hfill
    \subfloat[Three-vessel encounter scenario]{
        \includegraphics[width=\textwidth]{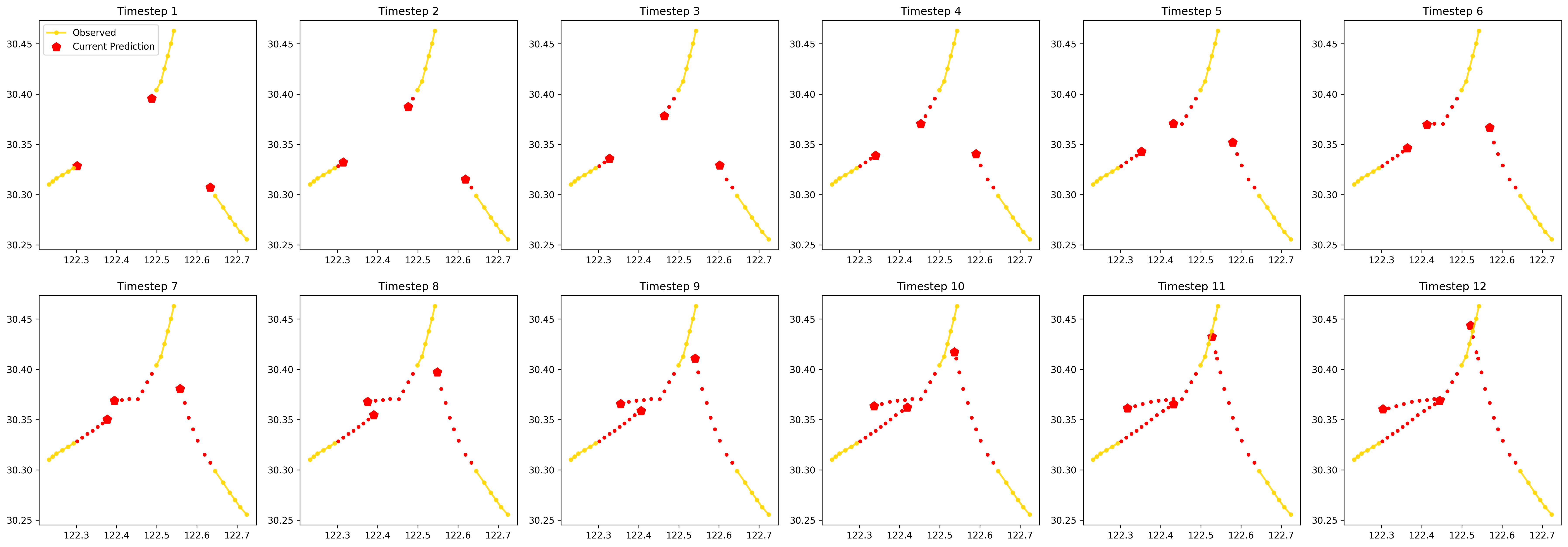}
        \label{fig:three_vessel_timesteps}
    }
    \caption{
    Predictions at different time steps. Yellow lines represent observed trajectories, while the red points show predictions at each time step. Subplots' timestep increases from left to right and top to bottom.
    }
    \label{fig_encounter_pred}
\end{figure*}

In contrast, the three-vessel encounter case (Fig.~\ref{fig_sustained}(c)) introduces greater complexity in interaction dynamics. The sustained intention probabilities become more evenly distributed (e.g., 0.4234, 0.4189), reflecting increased ambiguity and uncertainty in intention matching. This highlights the inherent difficulty deterministic prediction methods face in such complex scenarios, where a deterministic prediction is insufficient to represent diverse potential future modals.
  
To demonstrate DI-MTP's capability in capturing interactions between sustained and transient intentions, Fig.~\ref{fig_encounter_pred}(a) and \ref{fig_encounter_pred}(b) shows predictions for two-vessel and three-vessel encounter scenarios, respectively, showing optimal predictions in each case. In Fig.~\ref{fig_encounter_pred}(a), the vessels converge as time progresses. At the third time step, the upper-right vessel initiates a left turn and decelerates. At the fourth time step, the lower-left vessel turns left. By the fifth time step, both vessels have successfully avoided collision and passed each other. In Fig.~\ref{fig_encounter_pred}(b), all three vessels initially converge toward the center. The lower-right vessel begins early avoidance operations, while the distance between the upper and lower-left vessels decreases. The decreased point interval of the lower-left vessel indicates reduced speed, and at the fifth time step, the upper vessel operates a sharp right turn, successfully avoiding collision. These results demonstrate DI-MTP's ability to capture vessel encounters, maintain consistency in sustained-transient intention interactions, and generate globally consistent predictions across complex scenarios.

\subsubsection{Multimodality Analysis}
\label{multimodality ana}

\begin{figure*}[!t]  
   \centering
    \subfloat[DI-MTP multimodal predictions]{
        \includegraphics[width=\textwidth]{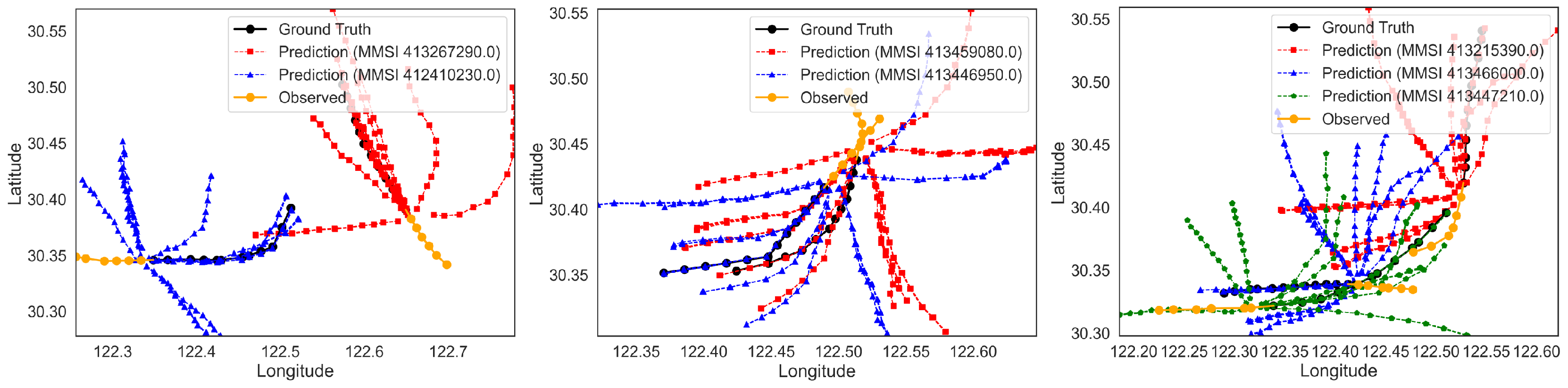}
    }
    
    \hfill
    \subfloat[GRU-CVAE multimodal predictions]{
        \includegraphics[width=\textwidth]{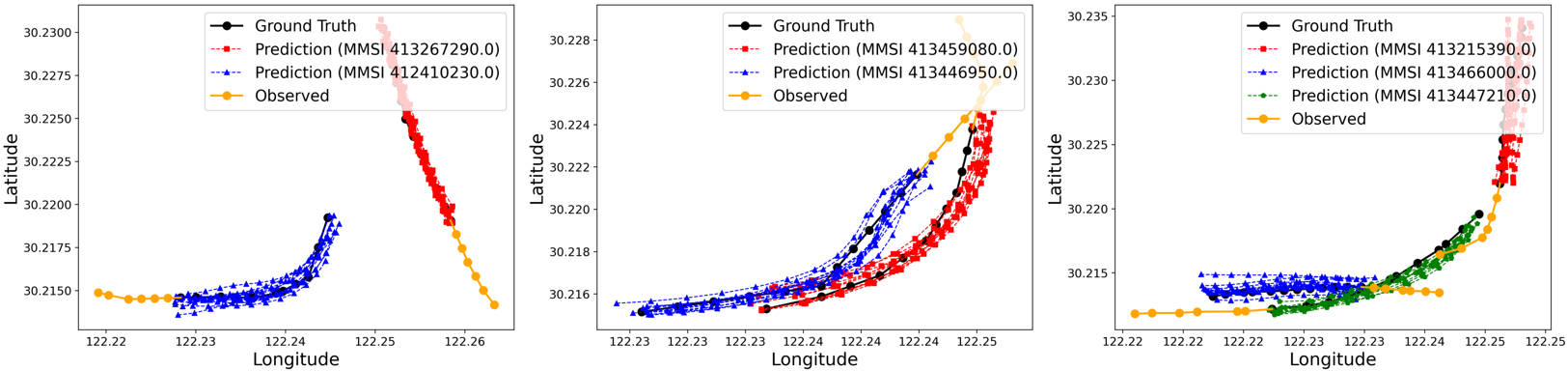}
    }
    
    \caption{
    Multimodal predictions in diverse scenarios. Yellow lines represent observed trajectories, black lines show true future trajectories and red/green/blue lines indicate multimodal predictions.
    }
    \label{fig_multimodality}
\end{figure*}

To highlight the distinction between modeling \emph{multimodality} and \emph{uncertainty}, we compare the prediction results of DI-MTP and GRU-CVAE across three representative scenarios: a non-encounter case, a two-vessel encounter, and a three-vessel encounter, as illustrated in Fig.~\ref{fig_multimodality}(a) shows the multimodal predictions generated by DI-MTP, (b) presents the results from GRU-CVAE.

In the non-encounter and three-vessel encounter cases, GRU-CVAE tends to generate predictions clustered around a single or a few similar modals, reflecting their inability to produce diverse future trajectories. In contrast, DI-MTP outputs distinct and well-separated modals, effectively representing the range of plausible futures.

Notably, in the two-vessel encounter scenario, GRU-CVAE outputs deviate from the ground truth, likely because the trajectory lies at the edge of the learned latent distribution, resulting in increased uncertainty and less reliable predictions.

Overall, this demonstrates the superiority of DI-MTP’s sustained-transient intentions in multimodal representation and robustness, validating its effectiveness for maritime trajectory prediction.



\section{Conclusion}
This paper presents DI-MTP, a unified model for multimodal vessel trajectory prediction that jointly models sustained and transient intentions. By capturing their interactions, DI-MTP improves prediction accuracy, diversity, and explainability in complex maritime scenarios. Experiments on real-world AIS datasets show that DI-MTP significantly reduces both ADE and FDE compared to state-of-the-art methods, and performs robustly across various encounter types, demonstrating strong adaptability to real-world navigation dynamics.

In future work, we plan to improve DI-MTP's interpretability and alignment by incorporating large language models (LLMs) for human-AI collaborative reasoning. LLMs will help generate explanations for intention modeling, validate decisions against COLREGs, and align multi-source inputs under environmental uncertainty.

\end{document}